\pdfoutput=1

\documentclass[11pt]{article}

\usepackage{acl}

\usepackage[T1]{fontenc}

\usepackage[utf8]{inputenc}

\usepackage{times}
\usepackage{latexsym}
\usepackage{microtype}
\usepackage{multirow}
\usepackage{graphicx}
\usepackage{tabu}
\usepackage{enumitem}
\usepackage{amsmath}
\usepackage{booktabs}
\usepackage{hyperref}
\usepackage{amssymb}
\usepackage{pifont}
\usepackage{microtype}
\usepackage{soul}
\usepackage{hyperref}

\title{In-BoXBART: Get Instructions into Biomedical Multi-Task Learning}

\author{Mihir Parmar$^1$ \hspace{8mm} Swaroop Mishra$^1$ \hspace{8mm} Mirali Purohit \hspace{8mm} Man Luo$^1$ \vspace{2mm} \\ \textbf{M. Hassan Murad}$^2$ \hspace{8mm} \textbf{Chitta Baral}$^1$ \vspace{4mm} \\ $^1$Arizona State University, USA \\ $^2$Mayo Clinic, USA}

\begin{document}
\maketitle
\begin{abstract}

Single-task models have proven pivotal in solving specific tasks; however, they have limitations in real-world applications where multi-tasking is necessary and domain shifts are exhibited. Recently, instructional prompts have shown significant improvement towards multi-task generalization; however, the effect of instructional prompts and Multi-Task Learning (MTL) has not been systematically studied in the biomedical domain. Motivated by this, this paper explores the impact of instructional prompts for biomedical MTL. We introduce the BoX, a collection of 32 instruction tasks for \textbf{B}i\textbf{o}medical NLP across (\textbf{X}) various categories. Using this meta-dataset, we propose a unified model termed as In-BoXBART, that can jointly learn all tasks of the BoX without any task-specific modules. To the best of our knowledge, this is the first attempt to propose a unified model in the biomedical domain and use instructions to achieve generalization across several biomedical tasks. Experimental results indicate that the proposed model: 1) outperforms single-task baseline by $\sim$3\% and multi-task (without instruction) baseline by $\sim$18\% on an average, and 2) shows $\sim$23\% improvement compared to single-task baseline in few-shot learning (i.e., 32 instances per task) on an average. Our analysis indicates that there is significant room for improvement across tasks in the BoX, implying the scope for future research direction.\footnote{\scriptsize\url{https://github.com/Mihir3009/In-BoXBART}}

\end{abstract}

\section{Introduction}

\begin{figure}[t]
    \centering
    \includegraphics[width=\linewidth]{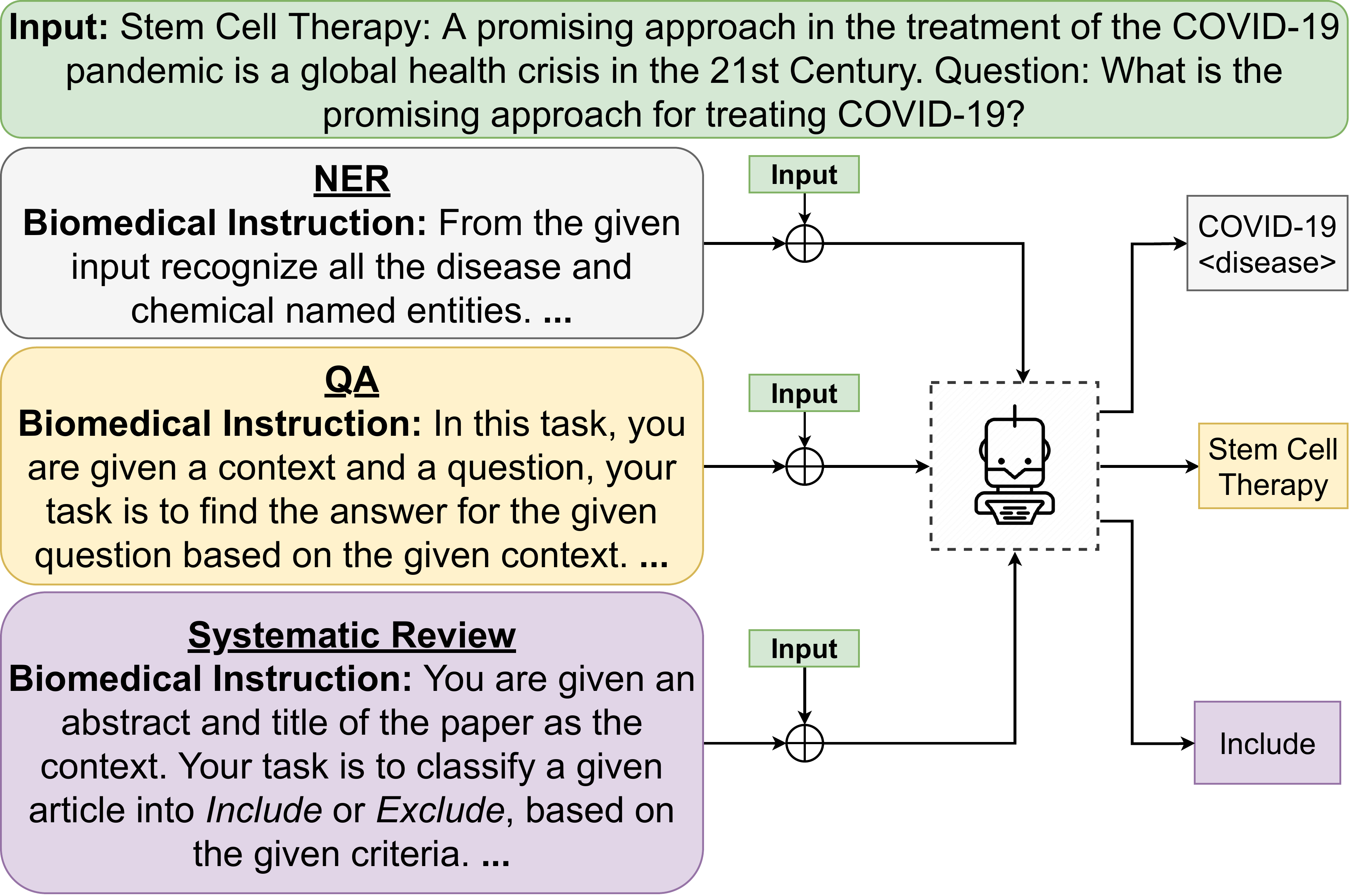}
    \caption{Schematic representation of multi-tasking in biomedical domain using instructional prompts. In this approach, a model is allowed to utilize tasks to get familiar with instructions and use them to map a given input to its corresponding output.}
    \label{fig:intro_fig}
\end{figure}

For long, task-specific models have played a central role in achieving state-of-the-art performance in both general and biomedical NLP \cite{wang2021pre, banerjee2021biomedical}. During 2017-2019, pre-train and fine-tune paradigm \cite{liu2021pre} became the prevalent approach in NLP. Due to success of Language Models (LMs) in the biomedical domain such as BioBERT \cite{lee2020biobert}, ClinicalXLNET \cite{huang2019clinical}, and others \cite{alrowili2021biom, kraljevic2021medgpt, phan2021scifive}, this paradigm is widely used for creating many task-specific models \cite{wang2021pre, banerjee2021biomedical}. However, task-specific models have limitations to real-world applications because this approach is computationally expensive (i.e., require large computational resources) and time-consuming \cite{strubell2019energy, schwartz2020green}. Hence, there is a need for generalization where a single model can perform various tasks leading to a computationally efficient approach. Past attempts have been made in general-domain NLP to achieve generalization across tasks such as MQAN \cite{mccann2018natural}, UNICORN \cite{lourie2021unicorn}, and UnifiedQA \cite{khashabi2020unifiedqa}. However, approaches to achieve generalization across various biomedical NLP tasks have not been systematically studied. Hence, this paper studies the multi-tasking approach that can generalize over different biomedical NLP tasks. Figure \ref{fig:intro_fig} shows the overview of our proposed multi-tasking approach where the single model can perform various biomedical NLP tasks.


Recently, prompt-based models have been widely used because of their ability to achieve generalization instead of task-specific models \cite{liu2021pre}. \citet{mishra2021cross, wei2021finetuned} and \citet{sanh2021multitask} show the effectiveness of instructional prompts in generalizing on seen as well as unseen general-domain NLP tasks. In this paper, we adapt this instructional prompt-based approach for the first time to achieve generalization across various biomedical NLP tasks. To this extent, this paper introduces a collection of 32 instruction tasks for \textbf{B}i\textbf{o}medical NLP across (\textbf{X}) various categories (BoX) and proposes a unified model that can generalize over 32 different biomedical NLP tasks. The proposed unified model (i.e., In-BoXBART) is trained on the instruction-based meta-dataset (i.e., BoX) and evaluated on each task individually from the BoX.

To evaluate the proposed approach, we compare our model (i.e., In-BoXBART) with two baselines: (1) single-task models (i.e., models trained on one task and evaluated on the same task), and (2) multi-task model (i.e., a single model trained on a combination of all tasks) without instructions. Experimental results show that In-BoXBART outperforms single-task baseline by $\sim$3\%, and multi-task baseline by $\sim$18\%. We also analyze few-shot learning scenario using In-BoXBART since obtaining annotated data in the biomedical domain is costly and time-consuming~\cite{luo2022improving}. In the few-shot setting (i.e., 32 instances per task), In-BoXBART outperforms the single-task baseline by 23.33\%. This indicates that Multi-Task Learning (MTL) and instruction-tuning have an advantage in the low resources settings. Although the performance of the In-BoxBART is promising, our analysis reveals that there is still room for improvement on some tasks, implying the scope for future research direction. Concisely, our contributions can be summarized in three folds:

\begin{enumerate}[nosep,noitemsep,leftmargin=*]
    \item This paper introduces the first benchmark meta-dataset in biomedical domain, i.e., BoX: a collection of 32 instruction tasks for Biomedical NLP across (X) various categories. Each task is processed in a unified format and equipped with instructions that can be used to train sequence-to-sequence models.
    \item Using this meta-dataset, we propose an instruction-tuned Bidirectional and Auto-Regressive Transformer (BART) model, termed as In-BoXBART. The comparison of In-BoxBART and two baselines shows that In-BoXBART outperforms single-task baseline by $\sim3\%$ and multi-task (without instruction) baseline by $\sim18\%$.
    \item In the few-shot setting, we show that In-BoxBART significantly outperforms the single-task baseline by $\sim23\%$. This indicates the potential application of instruction-tuning in the biomedical domain where annotated data is difficult to obtain.
\end{enumerate}


\section{Related Work}
\paragraph{Multi-task Learning}
Owing to the problems associated with single-task learning in terms of their space and time requirements, several multi-task learning approaches have been proposed over the years. DecaNLP~\cite{mccann2018natural} built a multi-tasking model by converting format of each tasks to question answering format. Several other works have followed similar approach, for example,  by converting tasks to reading comprehension ~\cite{Mishra2020TowardsQF} and textual entailment format ~\cite{wang2021entailment}. The multitasking model T5~\cite{raffel2020exploring} was built with the help of a unified framework that converts all text-based language problems into a text-to-text format. SCIFIVE~\cite{phan2021scifive} involved building a text-to-text model for the biomedical literature. \citet{aghajanyan2021muppet} introduced pre-finetuning, an additional large-scale learning stage between language model pre-training and fine-tuning to improve multitask learning performance. Models empowered by multi-task learning have achieved SOTA in many different tasks, e.g., Question Answering (QA)~\cite{khashabi2020unifiedqa}, commonsense reasoning~\cite{lourie2021unicorn} and structured knowledge grounding tasks~\cite{xie2022unifiedskg}.


\paragraph{Instruction Learning}
The turking test~\cite{efrat2020turking} was proposed to measure the efficacy of models to follow instructions. Studies have been made to investigate the effect of natural language instructions on model performance \cite{hase2021can, ye2021zero, zhong2021adapting, weller2020learning}. Moreover, \citet{mishra2021cross} proposed Natural Instructions which break down each task to multiple sub-tasks that help models in following instructions and subsequently generalize to unseen tasks (i.e., cross-task generalization). FLAN~\cite{wei2021finetuned} and T0~\cite{sanh2021multitask} models were built by leveraging instruction/prompt-tuning on diverse range of tasks and achieving zero-shot generalization on target unseen tasks. Task reframing~\cite{mishra2021reframing} proposed several guidelines to reframe task instructions to improve model response to follow instructions. Analysis introduced to understand in-context learning better on a large set of training tasks \cite{min2021metaicl, min2022rethinking}. InstructGPT model \cite{ouyang2022training} is proposed, which is fine-tuned with human feedback to follow natural instructions. Furthermore, many works focused on investigating whether LMs understands meaning of natural language and prompts \cite{webson2021prompt, zhao2021ethical}. \citet{weller2020learning} and \citet{ye2021learning} use task descriptions to achieve generalization to new tasks. \citet{puri2022many} introduced instruction augmentation to improve model performance and sample complexity. \citet{wang2022instructionner} has developed instruction-based multi-task framework for few-shot Named Entity Recognition (NER) task. \citet{prasad2022grips} introduced Gradient-free Instructional Prompt Search (GrIPS) for improving task instructions for large LMs. Recently, many approaches have been proposed to improve model performance using instructions \cite{wu2021ai, wu2022promptchainer, lin2021few, kuznia2022less}.

\begin{figure}[t]
    \centering
    \includegraphics[width=\linewidth]{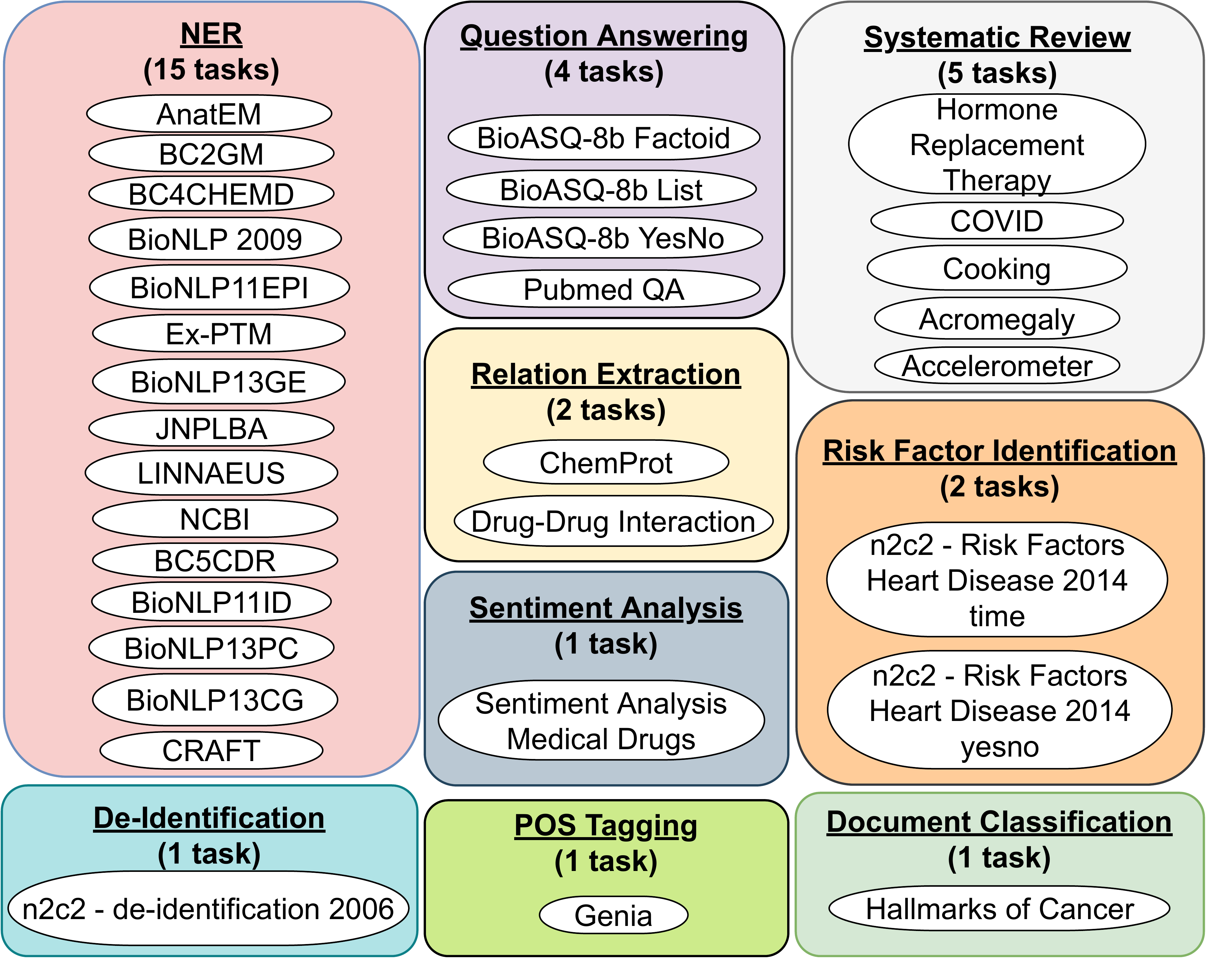}
    \caption{Schematic representation of 9 categories of tasks: each block represents one category with various tasks equipped with instruction.}
    \label{fig:dataset}
\end{figure}

\section{BoX}

We use 29 existing, widely adopted biomedical NLP datasets collected from various challenges, platforms and organizations to create BoX. We define the BoX as a benchmark dataset for biomedical MTL across 9 different categories. In the BoX, we reframed all the datasets as text generation tasks (see examples in Appendix \ref{app:examples}) and created 32 instruction tasks. BoX consists of high-quality human-authored Biomedical Instructions (BIs) for all 32 tasks. Figure \ref{fig:dataset} shows the 9 different categories and corresponding generated tasks. Each category is defined as colored box and each box contains instruction tasks re-purposed from original datasets.

\subsection{Tasks}

\label{sec:tasks}

\begin{table}[!t]
\centering
\resizebox{0.95\linewidth}{!}{
\begin{tabular}{lc} 
 \toprule
 Category & \# of training samples \\ [0.5ex] 
 \midrule
 NER & 82503 \\  
 De-identification & 106 \\ 
 POS Tagging & 16323 \\ 
 QA & 5778 \\
 RE & 23359 \\
 Sentiment Analysis & 2860 \\
 Systematic Review & 5761 \\
 Document Classification & 3119 \\
 Risk Factor Identification & 986 \\
 \midrule
 Total & 140795 \\
 \bottomrule
\end{tabular}
}
\caption{Size of training samples in each category}
\label{tab:category_statistics}
\end{table}

Table \ref{tab:category_statistics} shows the number of training samples we have used for each category. Further details of each instruction task statistics is shown in Appendix \ref{app:datatset_stats}. Each category and corresponding tasks from the BoX are defined as below:

\paragraph{Named Entity Recognition (NER)}
NER has been considered a necessary first step in processing literature for biomedical text mining where the model helps in identifying named entities such as protein, gene, chemical, disease, treatment. We use fifteen publicly available biomedical NER datasets \cite{crichton2017neural} to create instruction tasks.

\paragraph{De-Identification (DI)}
In this task, the model takes medical discharge records of a patient as input and identify Private Health Information (PHI) such as organizations, persons, locations, dates. We use n2c2 2006 de-identification challenge dataset \cite{uzuner2007evaluating} to perform this task.

\paragraph{Part-Of-Speech (POS) Tagging}
The goal of this task is to identify various POS tags from the biomedical text. We use GENIA corpus \cite{tateisi2005syntax} built from MEDLINE abstracts for the POS tagging task.

\paragraph{Question-Answering (QA)}
QA models receive a question and a corresponding context as input and output the relevant answer from the given context. To execute this task, we used the BioASQ-8b dataset \cite{nentidis2020overview} for different question types, i.e., yes/no, factoid, and list type questions. We created three different tasks from this dataset. Also, we use PubMedQA dataset \cite{jin2019pubmedqa} for this task.

\paragraph{Relation Extraction (RE)}
We used two datasets for this task: (1) CHEMPROT corpus from biocreative VI precision medicine track \cite{islamaj2019overview}, and (2) Drug-Drug Interaction (DDI) corpus from SemEval 2013 DDI Extraction challenge \cite{herrero2013ddi}. Here, we only consider binary RE tasks without any label describing the type of the relation. 

\paragraph{Systematic Review (SR)}
We have included data from the following five Systematic Reviews (SRs) that were conducted using the traditional (manual) process and published in relevant venues by Mayo Clinic physicians: (1) Hormone Replacement Therapy (HRT), (2) Cooking, (3) Accelerometer, (4) Acromegaly, and (5) COVID for this task \cite{parmar2021automation}. More details about these datasets creation and statistics are given in Appendix \ref{app:sr_datasets}.

\paragraph{Sentiment Analysis (SA)}
Analyzing the sentiment of people towards medical drugs is an essential task in the biomedical domain. To that effect, we use medical drug sentiment analysis dataset\footnote{\url{https://www.kaggle.com/arbazkhan971/analyticvidhyadatasetsentiment}} to identify one of three sentiments: (1) positive, (2) negative, and (3) neutral.

\paragraph{Document Classification (DC)}
We have used the Hallmarks of Cancer (HoC) dataset \cite{baker2016automatic} for this task.

\paragraph{Risk Factor Identification (RFI)}
The goal of this task is to identify risk factors for Coronary Artery Disease (CAD) in diabetic patients over time. For this, we used n2c2 2014 shared task track 2 dataset \cite{kumar2015creation} with two different purposes: (1) identify if the risk factor is presented in the medical discharge summary and (2) time of risk factor present in the discharge records.

\subsection{Biomedical Instructions}

\begin{figure}[t]
    \centering
    \includegraphics[width=\linewidth]{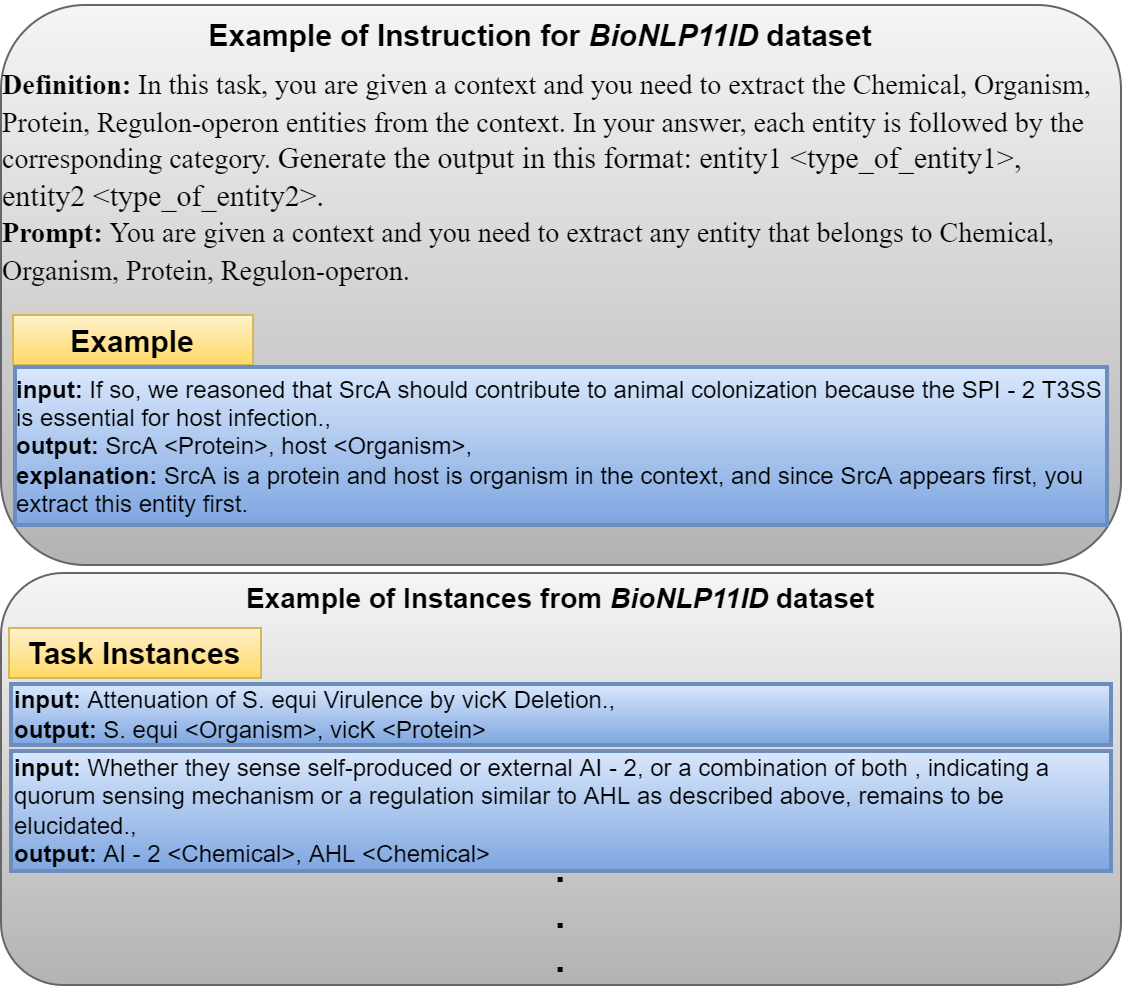}
    \caption{Example of Biomedical Instruction (BI) and task instances from \textbf{\textit{BioNLP11ID}} (NER) dataset.}
    \label{fig:example_instance}
\end{figure}

Motivated by \citet{mishra2021cross}, we have used a similar approach to create Biomedical Instructions (BIs). BI consists of natural language instructions that describe a task and contain instances of that task. Here, we introduce a unified schema to present BI and described how we can construct BI for each task given in the BoX. Figure \ref{fig:schema} illustrates the schematic representation of the schema, and Figure \ref{fig:example_instance} shows an example of BI that describes a ``Named Entity Recognition (NER)'' task accompanied with a few positive examples.

\subsubsection{Unified Schema}

\begin{figure}[t]
    \centering
    \includegraphics[width=0.7\linewidth]{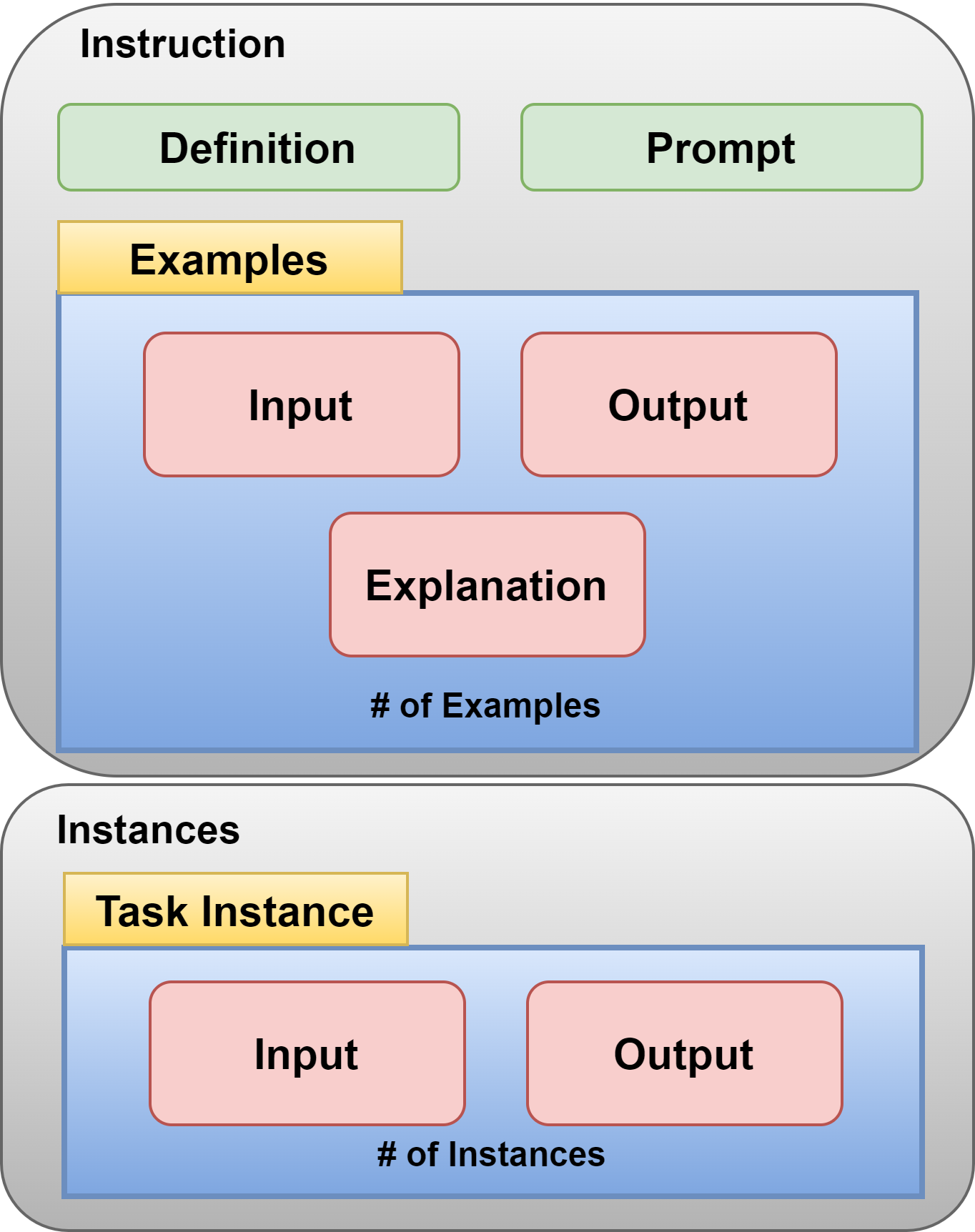}
    \caption{Unified schema used to create a Biomedical Instruction (BI).}
    \label{fig:schema}
\end{figure}

All BIs are mapped to the unified schema. As shown in Figure \ref{fig:schema}, unified schema consists of a definition, prompt, and positive examples. This schema helps in organizing each BI. Each of the elements of the schema is explained below:

\paragraph{Definition} contains the core explanation about the task and detailed instruction to the model that what needs to be done in the given task.

\paragraph{Prompt} is the short explanation of the task that needs to be done.

\paragraph{Examples} contain the input/output pairs of the task instance along with the explanation of how the output is generated. Generally, we provide 2-3 examples for each task.

\paragraph{Instances} contain the input/output pairs of training samples from the task datasets.

\subsubsection{Construction of BI}

We have created a BI for each dataset given in the BoX. To create BI, we manually fill in the fields of unified instruction schema (Figure \ref{fig:schema}). For each dataset, the BI is created by one author and were verified by other authors.

\paragraph{Quality of BIs} In the instruction verification process, we edit BIs if needed in terms of grammar, typos, ambiguity, etc. to improve the quality. According to \cite{beltagy2020longformer}, concise instructions are more beneficial compared to repetition, hence, we also redact repetition from BIs. As a result, our BIs consists of high-quality, short, and meaningful task definition, and prompts.

\paragraph{Positive examples and its explanation} For each dataset, we have provided 2-3 positive examples and corresponding explanations to give an idea of how to perform the given task. As we know, the selection of examples has an impact on model performance \cite{lu2021fantastically}. To that extent, we have been careful in selecting examples for text generation and classification tasks. For text generation, we have provided 2-3 examples with a detailed explanation about how the output is generated. For text classification tasks, we have included examples corresponding to each class with an explanation of why the particular class is assigned to a given input instance. All positive examples are drawn from training instances and have been removed from training in order to avoid repetition. All the explanations of examples pass through the verification process to maintain high quality.

\paragraph{Collection of input/output instances}
Since each biomedical NLP dataset included in the BoX has its own annotated input/output instances, we converted them into text-to-text format \cite{lourie2021unicorn}. Example of instances converted for each task is given in Appendix \ref{app:examples}. After this, we appended all instances tuple (i.e., <input, output>) with instruction schema (as shown in Figure \ref{fig:schema}).

\section{Problem Setup and Models}


\subsection{Problem setup}

Let us assume, we have input/output instances pair $(X_t, Y_t)$ for given task $t$. Along with that, each task is described in terms of its instruction ${BI}_t$.

\paragraph{Single-task models}
Traditional supervised models learn a mapping function ($f_M$) between input ($x$) and output ($y$), where $(x,y) \in ({X_t}^{\text{train}}, {Y_t}^{\text{train}})$ and are evaluated on the same task $({X_t}^{\text{test}}, {Y_t}^{\text{test}})$. We refer this setup as single-task learning.

\paragraph{Multi-task models}
In this setup, we combined training data and corresponding biomedical instruction of all tasks together. The goal of multi-task learning models is to learn mapping function ($f_M$) between input ($x$), output ($y$) and biomedical instruction ${BI}_t$, i.e., $f_M({BI}_t,x) = y$, where $(x,y) \in (X_t, Y_t)$. This model is evaluated on task-specific instances $(x,y) \in ({X_t}^{\text{test}}, {Y_t}^{\text{test}})$ In contrast to single-task models, a single model is used here to solve various tasks, hence, achieving generalization. We refer this setup as MTL.

\subsection{Models}

We propose an instruction-based model to achieve multi-tasking and compare it with two baselines: (1) single-task models, and (2) multi-task models without instructions. We have fine-tuned the BART (base) model \cite{lewis2019bart} to build baselines as well as the proposed model.

\subsubsection{Baselines}

\paragraph{Single-Task models}
As formulated in the single-task problem setup, we have trained the BART model on each task from the BoX and evaluated it on the same task.

\paragraph{Multi-task without instruction}
To build this baseline, we have combined training data of each task from the BoX together without appending BIs and trained a single model on the combined data. We refer this model as Vanilla-BoXBART. This model is evaluated on each task of the BoX.

\subsubsection{Proposed Model}

As formulated in the multi-task problem setup, we have combined training data and the corresponding BI of each task. To combine instruction with input instances, we map a BI and an input ($x$) into the textual format and obtain $enc({BI}_t, x)$. After that, BART model is used to predict an output ($y$) using a mapping function $f_M: enc({BI}_t, x) \to y$. To perform encoding, a standard NLP paradigm of mapping is used, i.e., mapping an input to text. Here, we map each element of BI (i.e., definition and positive examples as shown in the schema) to a textual format and append it before the input instances. After appending BI of each task to instances, we combined all training data of each task. Now, we fine-tuned the BART model with this combined instruction meta-dataset. We refer this instruction-tuned model as In-BoXBART.

\section{Experiments and Analysis}

\subsection{Experimental Setup}
\label{subsec:ex_setup}

We have used BART (base) model to build all baselines and proposed model. All the experiments are performed using \textit{Quadro RTX 8000} GPU. All models are trained for 3 epochs. In particular, we have used \textit{huggingface implementation} \cite{wolf2019huggingface} of the BART and its pre-defined functions for the training and evaluation with default parameters.



\paragraph{Instance Selection} As we know, BART (base) can accept the input of a maximum $1024$ token length. Since there are few instances in some datasets that exceed this limit (after including instructions), we have discarded those instances while creating instruction tasks. We have also removed the same instances while training two baselines to do a fair comparison. We have discarded long samples (>1024 token length) from validation and testing data as well.


\paragraph{Example Selection} As discussed in \citet{lu2021fantastically}, the selection and order of the examples included in instructions matters for mainly classification tasks and affects the performance of the model. We empirically conclude that the proposed model benefits from ignoring examples from biomedical instructions for classification tasks during training and evaluation. Hence, we have discarded all examples from the BIs associated with the classification instruction tasks.


\paragraph{Instance Sampling} Some classification datasets used to create the BoX are imbalanced. To balance these datasets, we have applied the sampling techniques \cite{poolsawad2014balancing} before using datasets to create BoX. In particular, we have analyzed three sampling techniques: (1) under-sampling, (2) average-sampling, and (3) over-sampling. In under-sampling, we have reduced instances for all the classes to the class with the lowest number of instances. In contrast, we have over-sampled instances via replication of random instances to the class with the highest number of instances to achieve over-sampling. In average sampling, we calculated mean of number of instances across all the classes and over-sampled or under-sampled instances accordingly for each class.

\paragraph{Few-shot setting} Similar to the \cite{schick2020s}, we have started with 32 randomly selected instances for each instruction task from the BoX to exhibit few-shot learning. After that, we have increased randomly selected instance instances per task to 100/$1k$/$4k$. If any task have already less number of instances than the threshold (i.e., 100/$1k$/$4k$), we keep all the instances from that task. While selecting the instances, we made sure that we select balanced data for the classification tasks. Moreover, the BoX contains an average $6k$ instances per task.

\paragraph{Evaluation Metric} We use Rouge-L \cite{lin2004rouge} as our evaluation metric since we treat all the tasks as text generation problems. We also use $F_1$-Score for evaluations.

\begin{table*}[t!]
\centering
\small
\resizebox{0.95\linewidth}{!}{

\begin{tabular}{@{}llcccccc@{}}
\toprule
\multirow{2}{*}{Category} & \multirow{2}{*}{Task} & \multicolumn{3}{c}{Rouge-L} & \multicolumn{3}{c}{$F_1$-Score} \\
\cmidrule(lr){3-5} \cmidrule(lr){6-8}
~ & ~ & Single Task & V-BB & I-BB & Single Task & V-BB & I-BB \\
 \toprule
 \multirow{16}{*}{NER}
 & AnatEM & \textbf{84.88} & 32.30 & 83.93 & \textbf{85.55} & 33.50 & 84.61  \\
 & BC2GM & \textbf{77.66} & 50.87 & 74.10 & \textbf{78.56} & 50.86 & 75.03 \\
 & BC4CHEMD & \textbf{88.85} & 71.05 & 86.50 & \textbf{89.06} & 71.44 & 86.97\\
 & BC5CDR & \textbf{74.83} & 69.81 & 74.76 & 75.13 & 70.11 & \textbf{75.24} \\
 & BioNLP11EPI & 84.64 & 50.10 & \textbf{87.60} & 84.95 & 52.85 & \textbf{88.04} \\
 & BioNLP11ID & 71.08 & 59.12 & \textbf{72.64} & 71.64 & 60.15 & \textbf{73.39} \\
 & BioNLP13CG & 64.19 & 55.18 & \textbf{67.72} & 61.68 & 53.88 & \textbf{65.09} \\
 & BioNLP13GE & 83.74 & 49.30 & \textbf{86.71} & 84.08 & 51.78 & \textbf{87.39} \\
 & BioNLP13PC & 70.42 & 53.06 & \textbf{72.46} & 66.89 & 51.61 & \textbf{67.77} \\
 & BioNLP09 & 85.16 & 51.54 & \textbf{88.09} & 85.54 & 54.31 & \textbf{88.48} \\
 & CRAFT & 63.72 & 51.85 & \textbf{64.10} & 63.92 & 52.31 & \textbf{64.30} \\
 & Ex-PTM & 82.32 & 49.61 & \textbf{83.73} & 82.38 & 52.07 & \textbf{84.49} \\
 & JNLPBA & \textbf{71.65} & 69.37 & 71.54 & \textbf{70.79} & 68.60 & 70.26 \\
 & NCBI & \textbf{89.51} & 74.46 & 86.11 & \textbf{89.81} & 75.55 & 80.91 \\
 & linnaeus & \textbf{94.43} & 44.99 & 93.46 & 93.21 & 44.59 & \textbf{93.77} \\
 & ---------------------------- & -------- & -------- & -------- & -------- & -------- & -------- \\
 & Average & 79.14 & 55.51 & \textbf{79.54} & 78.88 & 56.24 & \textbf{79.45} \\
 \midrule
 \multirow{1}{*}{DI}
 & DI 2006 & 12.60 & 46.38 & \textbf{50.82} & 10.60 & 43.28 & \textbf{47.45} \\
 \midrule
 \multirow{1}{*}{POS}
 & Genia & \textbf{71.45} & 27.94 & 71.26 & 70.48 & 27.50 & \textbf{71.99} \\
 \midrule
 \multirow{4}{*}{QA}
 & BioASQ8b (factoid) & \textbf{52.95} & 51.14 & 47.28 & \textbf{54.67} & 53.52 & 49.51 \\
 & BioASQ8b (list) & \textbf{38.96} & 19.87 & 36.11 & - & 17.74 & 35.59 \\
 & BioASQ8b (yesno) & 61.74 & 62.61 & \textbf{68.25} & 63.48 & 62.61 & \textbf{68.25} \\
 & PubMedQA & \textbf{27.12} & 25.48 & 24.49 & \textbf{31.44} & 30.74 & 29.58 \\
 & ---------------------------- & -------- & -------- & -------- & -------- & -------- & -------- \\
 & Average & \textbf{45.19} & 39.78 & 44.03 & \textbf{46.39} & 41.15 & 45.73 \\
 \midrule
 \multirow{2}{*}{RE}
 & ChemProt & 76.08 & 76.00 &\textbf{81.61} & \textbf{63.89} & 52.17 & 63.22 \\
 & DDI & \textbf{91.78} & 82.97 & 89.35 & \textbf{94.10} & 82.97 & 89.35 \\
 & ---------------------------- & -------- & -------- & -------- & -------- & -------- & -------- \\
 & Average & 83.04 & 79.48 & \textbf{85.48} & \textbf{79.00} & 67.57 & 76.28\\
 \midrule
 \multirow{1}{*}{SA}
 & Medical Drugs & \textbf{47.51} & 46.39 & 47.37 & \textbf{47.51} & 46.39 & 47.37 \\
 \midrule
 \multirow{6}{*}{SR}
 & Accelerometer & 74.65 & 72.54 & \textbf{81.25} & 74.65 & 72.54 & \textbf{81.25} \\
 & Acromegaly & 80.21 & \textbf{81.77} & 80.71 & 80.21 & \textbf{81.77} & 80.71 \\
 & COVID & 74.81 & 76.30 & \textbf{77.28} & 74.81 & 76.30 & \textbf{77.28} \\
 & Cooking & 71.71 & 82.93 & \textbf{83.25} & 71.71 & 82.93 & \textbf{83.25} \\
 & HRT & 75.68 & 77.17 & \textbf{82.70} & 75.68 & 77.17 & \textbf{82.70} \\
 & ---------------------------- & -------- & -------- & -------- & -------- & -------- & -------- \\
 & Average & 75.41 & 78.14 & \textbf{81.04} & 75.41 & 78.14 & \textbf{81.04} \\
 \midrule
 \multirow{1}{*}{DC}
 & HoC & \textbf{88.53} & 49.64 & 82.53 & \textbf{88.53} & 49.51 & 82.53 \\ 
 \midrule
 \multirow{3}{*}{RFI}
 & RFHD 2014 (yesno) & 57.21 & 64.97 & \textbf{69.17} & 57.21 & 64.97 & \textbf{69.17} \\ 
 & RFHD 2014 (time-riskfactor) & 66.18 & 0.97 & \textbf{85.24} & 66.18 & 0.97 & \textbf{85.28} \\
 & ---------------------------- & -------- & -------- & -------- & -------- & -------- & -------- \\
 & Average & 72.87 & 57.30 & \textbf{77.21} & 61.69 & 32.97 & \textbf{77.22} \\ 
 \midrule
 \multirow{1}{*}{Average}
 & - & 70.51 & 55.55 & \textbf{73.49} & 70.15 & 55.21 & \textbf{73.01} \\ 
 \bottomrule
\end{tabular}
}
\caption{Results comparison between single-task baseline, Vanilla-BoXBART and In-BoXBART in terms of Rouge-L and $F_1$-Score. All the results for $F_1$-Score are presented in \%. V-BB: Vanilla-BoXBART, I-BB: In-BoXBART, RFHD: Risk Factor for Heart Disease.}
\label{tab:main-results}
\end{table*}

\subsection{Results and Findings}

\paragraph{Effect of Sampling} As mentioned above, we conduct three experiments to analyze the effect of sampling on In-BoXBART. We train our model using training data obtained from (1) under-sampling, (2) average-sampling, and (3) over-sampling. We achieve on an average (across all instruction tasks) 69.62, 70.23 and 73.49 Rouge-L for under-, average- and over-sampling, respectively. Here, we observe from the experimental results that over-sampling gives better performance compared to under- and average-sampling since there is a loss of training data samples for under- and average-sampling. Hence, we report results of over-sampling as the main result in Table \ref{tab:main-results}.

\paragraph{Performance comparison} Table \ref{tab:main-results} presents the results for single-task model, Vanilla-BoXBART and In-BoXBART. We can see from Table \ref{tab:main-results} that the single-task model, Vanilla-BoXBART, and In-BoXBART achieve on an average (across all tasks) Rouge-L of 70.51, 55.55, and 73.49, respectively. They achieve 70.15\%, 55.21\%, and 73.01\% $F_1$-Score, respectively, exhibiting the same performance behaviour as Rouge-L. Hence, we use Rouge-L for further comparisons. From the result, we can observe that Vanilla-BoXBART reduces the complexity compared to the single-task model (i.e., 110 million parameters \textit{vs.} 32x110 million parameters), however, on an average the performance drops by 14.96\% in terms of Rouge-L, and  compared to single-task models. This indicates that multi-task learning in the biomedical domain is more difficult than general domain NLP since many previous works have shown that the multi-task model outperforms the single-task model \cite{lourie2021unicorn, mccann2018natural}. On the other hand, In-BoXBART, which has the same complexity as Vanilla-BoXBART, significantly outperforms Vanilla-BoXBART by on average 17.94\%, and also outperforms the single-task model by a 2.98\% margin, precisely. This indicates the benefit of using instructions to achieve the MTL in the biomedical domain. 

\begin{figure}[t!]
    \centering
    \includegraphics[width=\linewidth]{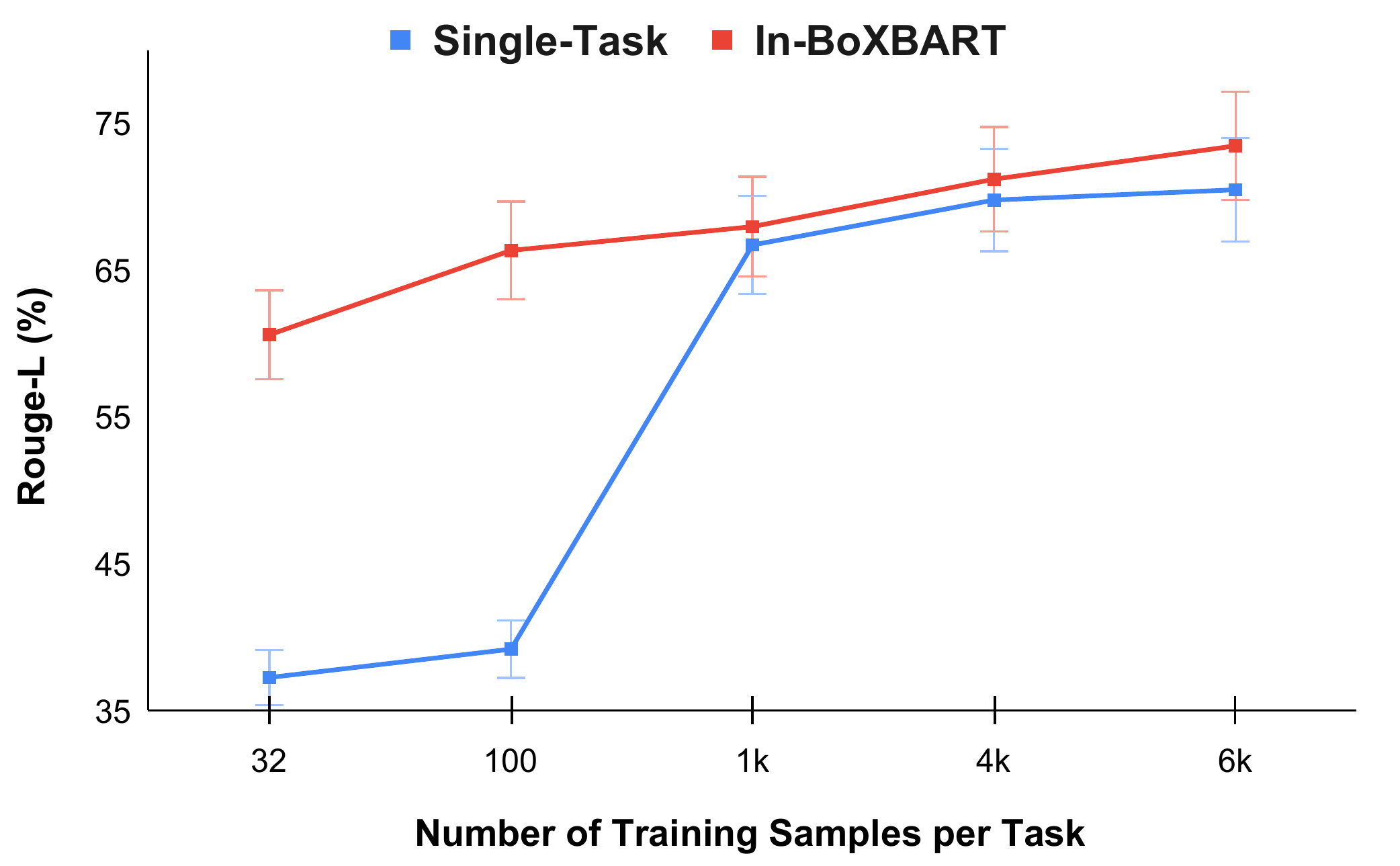}
    \caption{Comparison of on an average Rouge-L across all instruction tasks between single-task and In-BoXBART based on the average number of training instances per task.}
    \label{fig:fewshot_graph}
\end{figure}

\paragraph{Effect of instruction in few-shot learning} We have compared the average Rouge-L of In-BoXBART with a single-task baseline for few-shot setting. Figure \ref{fig:fewshot_graph} shows the relative performance of In-BoXBART compared to single-task baseline. We have shown results for all few-shot learning experiments in Appendix \ref{app:fewshot_results}. From the results, we see that In-BoXBART achieves on an average 60.64\% Rouge-L and the single-task model achieves 37.31\% for 32 instances per task. In-BoxBART significantly outperforms the single-task baseline by 23.33\%. From Figure \ref{fig:fewshot_graph}, we can see that In-BoXBART consistently perform better compared to the baseline. As we know, obtaining a large annotated dataset in the biomedical domain is difficult, time-consuming and costly. From few-shot learning, we can see that instructions are beneficial in achieving high performance compared to task-specific models.


\subsection{Analysis}

\paragraph{For which tasks, instruction is helpful?} From Table \ref{tab:main-results}, we can see that In-BoXBART outperforms baselines for 5 categories, i.e., NER, de-identification, RE, SR and risk factor identification. From this, we can see that instructions are more helpful in these five categories. However, In-BoXBART achieves performance lower or par with the single-task baseline for the tasks from QA, POS tagging, sentiment analysis and document classification which indicates room for improvement in this direction.

\paragraph{Which are harder tasks to solve using instructions?} Although instructions help in achieving better performance for some tasks compared to the single-task model, the overall performance is still lower. For example, instruction improves performance for de-identification, but overall performance on this task is only 50.82\% which can be improved. A similar pattern we can see for BioNLP12CG and CRAFT from NER; BioASQ-8b (factoid, list) and PubmedQA from QA; and Medical Drug from the sentiment analysis category. In general, we can observe that tasks that include either multi-class scenario or answer generation from the context are most likely to be harder to solve using instructions. For example, CRAFT and BioNLP13CG have 6 entity types which are higher than any other tasks from NER, and we can see that the performance for these two tasks is lower compared to other tasks of NER. 

\paragraph{For which tasks, instruction is the most beneficial in few shot setting?} From the results shown in Appendix \ref{app:fewshot_results}, tasks from the NER, de-identification, QA, sentiment analysis and risk factor identification shows on average larger improvement compared to baselines for the few-shot settings (i.e., 32 and 100 instances per task). This indicates that instructions are beneficial for the tasks from the above categories.

\section{Discussion}


\paragraph{Can we design better instructions?} Since instruction teach the model how to solve a given task, domain specific information rich instructions can improve model performance. One potential way is to use the knowledge of domain experts. However, designing a good biomedical instruction can be one research direction.


\paragraph{How to handle long-context input?} Training instances of many biomedical datasets consist Electronic Health Records (EHRs) or discharge summaries of patients. Because of this, these instances are long and exceed the maximum input length of LMs such as BERT, BART. In this scenario, encoding extra information in terms of prompts or instructions becomes difficult. One potential solution is to use Longformer \cite{beltagy2020longformer}, and another solution is to use T5 kind of models which use relative position embeddings so that the inference length can be longer~\cite{luo2022choose}.




\paragraph{How to handle multi-class classification tasks?} Multiple classes cause an issue while creating biomedical instructions because we can not present one example per class. If we do that, the encoding of BI and input will exceed the maximum length of LMs. A naive solution is to select examples of a few labels or remove the examples. However, this will cause a label bias issue or performance degradation. Potential future research direction can be designing a methodology to handle multi-class classification tasks.

\paragraph{How far we are from the SOTA?} We have presented preliminary comparison of our results w.r.t. state-of-the-art (SOTA) single-task systems for 21 instruction tasks from the BoX as shown in Appendix \ref{app:sota_results}. Form the results, we can see that the performance of the proposed model remains far from the SOTA for some tasks, indicating significant room for further research in this domain.

\section{Summary and Conclusions}

This research shows the impact of instructions in MTL for the first time in the biomedical domain. To this extent, we introduced the BoX, a first benchmark dataset consisting of 32 instruction tasks across various biomedical NLP domains. Using this meta-dataset, we proposed a unified model, i.e., In-BoXBART which outperforms single-task baseline and Vanilla-BoxBART by $\sim3\%$ and $\sim18\%$, respectively. Our proposed approach also shows an effective performance for a few-shot setting which is more beneficial in the biomedical domain where obtaining large annotated datasets is difficult. We hope that the BoX benchmark, In-BoXBART, and experimental results encourage future research into more unified models for biomedical NLP.

\section*{Acknowledgement}

The authors acknowledge support from ONR award number N00014-20-1-2332 for this project. The authors thank the anonymous reviewers for their feedback.

\bibliography{custom}
\bibliographystyle{acl_natbib}

\newpage

\appendix

\section{Statistics of Instruction Tasks}
\label{app:datatset_stats}

This section provides all the statistics of training, validation and inference data used for experiments in Table \ref{tab:dataset_stats}. All the number of instances provided in Table \ref{tab:dataset_stats} are calculated after discarding the instances with more than 1024 token length as described in the section \ref{subsec:ex_setup}. We have divided the dataset into standard 70/10/20 splits for train/validation/test if there is no separate validation and testing set provided in the dataset.

\begin{table*}[t]
\centering
\small
\resizebox{0.65\linewidth}{!}{

\begin{tabular}{@{}llccc@{}}
\toprule
\multirow{2}{*}{Category} & \multirow{2}{*}{Tasks} &  \multicolumn{3}{c}{\# of Instances}\\
\cmidrule(lr){3-5}
~& ~& Train & Dev & Test \\ 
 \toprule
 \multirow{16}{*}{NER}
 & AnatEM  & 3507 & 1121 & 2303 \\ 
 & BC2GM & 6427 & 1291 & 2570 \\ 
 & BC4CHEMD & 14466 & 14568 & 12397 \\ 
 & BC5CDR & 4940 & 4940 & 5158 \\
 & BioNLP11EPI & 3796 & 1242 & 2836 \\ 
 & BioNLP11ID & 2466 & 780 & 1869 \\ 
 & BioNLP13CG & 4591 & 1489 & 2759 \\ 
 & BioNLP13GE & 1503 & 1663 & 1937 \\
 & BioNLP13PC & 2945 & 1070 & 1997 \\ 
 & BioNLP09 & 4710 & 1013 & 1699 \\ 
 & CRAFT & 12839 & 4423 & 8882 \\ 
 & Ex-PTM  & 855 & 278 & 1160 \\
 & JNLPBA & 15124 & 1533 & 3152 \\ 
 & NCBI & 2922 & 488 & 538 \\ 
 & linnaeus & 1484 & 524 & 993 \\
 \midrule
 \multirow{1}{*}{DI}
 & DI 2006 & 106 & 22 & 27 \\ 
 \midrule
 \multirow{1}{*}{POS}
 & Genia & 16323 & 2174 & 2035 \\ 
 \midrule
 \multirow{2}{*}{QA}
 & BioASQ8b (factoid) & 695 & 16 & 115 \\ 
 & BioASQ8b (list) & 373 & 8 & 45 \\ 
 & BioASQ8b (yesno) & 543 & 16 & 115 \\ 
 & PubMedQA & 4167 & 500 & 473 \\
 \midrule
 \multirow{2}{*}{RE}
 & ChemProt & 3350 & 2415 & 2660 \\ 
 & DDI & 20009 & 2780 & 2660 \\
 \midrule
 \multirow{1}{*}{SA}
 & Medical Drugs & 2860 & 526 & 804 \\ 
 \midrule
 \multirow{6}{*}{SR}
 & Accelerometer & 499 & 58 & 142 \\ 
 & Acromegaly & 663 & 80 & 192 \\ 
 & COVID & 2385 & 300 & 675 \\ 
 & Cooking & 735 & 84 & 205 \\ 
 & HRT & 1479 & 171 & 410 \\
 \midrule
 \multirow{1}{*}{DC}
 & HoC & 3119 & 445 & 890 \\ 
 \midrule
 \multirow{3}{*}{RFI}
 & RFHD 2014 (yesno) & 834 & 360 & 451 \\ 
 & RFHD 2014 (time-riskfactor) & 152 & 177 & 69 \\
 \midrule
 \multirow{1}{*}{Total}
 & - & 140795 & 46554 & 64561 \\ 
 \bottomrule
\end{tabular}
}
\caption{Statistics of training (i.e., Train), validation (i.e, Dev) and evaluation (i.e., Test) data for all instruction tasks from the BoX. RFHD: Risk Factor for Heart Disease.}
\label{tab:dataset_stats}
\end{table*}

\section{Instruction Tasks and Examples}
\label{app:examples}


To build all the models (baselines, proposed model and few-shot learning), we adapt the unified format for all the tasks of BoX. We converted all the tasks into the text-to-text format, including the classification tasks. Table \ref{tab:examples} shows an example of input and output from each category. Moreover, we have also re-purposed some biomedical datasets to create more than one task as described in the section \ref{sec:tasks}.


\begin{table*}[t]
\centering
\small
\resizebox{0.95\linewidth}{!}{

\begin{tabular}{ p{2cm} | p{1.6cm} | p{6cm} | p{5cm} }


\toprule
\multirow{1}{*}{Category} & \multirow{1}{*}{Task} & \multirow{1}{*}{Input} & \multirow{1}{*}{Output}\\
\toprule
\multirow{3}{*}{NER} & \multirow{3}{*}{\shortstack{BC5CDR}} & Such interactions may result in serious cardiovascular complications even after cessation of an infusion of ritodrine. & \multirow{3}{*}{\shortstack{cardiovascular complications <Disease>, \\ ritodrine <Chemical>}} \\
\midrule
\multirow{11}{*}{de-identification} & \multirow{11}{*}{\shortstack{DI2006}} & 757085252 HLGMC 1228824 18705/6o5b 3/25/1993 12:00:00 AM CONGESTIVE HEART FAILURE . Unsigned DIS Report Status : Unsigned ADMISSION DATE : 3/25/93 DISCHARGE DATE : 4/4/93 PRINCIPAL DIAGNOSIS : congestive heart failure . ASSOCIATED DIAGNOSIS : aortic stenosis ; coronary artery disease , status post multi vessel coronary artery bypass graft surgery , ... , M.D. TR : go / bmot DD : 4/4/93 TD : 04/06/93 CC : [ report\_end ] & \multirow{11}{*}{\shortstack{3/25 <DATE>, 18705/6o5b <ID>,\\757085252 <ID>, go / bmot <DOCTOR>,\\4/4 <DATE>, 04/06 <DATE>}} \\
\midrule
\multirow{3}{*}{POS-Tagging} & \multirow{3}{*}{\shortstack{Genia}} & \multirow{3}{*}{Binding sites were mapped for each factor .} & Binding <VBG> sites <NNS> were <VBD> mapped <VBN> for <IN> each <DT> factor <NN> . <.> \\
\midrule
\multirow{8}{*}{QA} & \multirow{8}{*}{\shortstack{BioASQ8b\\(factoid)}} & Context: Hyperosmia is suspected in pregnancy; however, no empirical study using validated measures of olfactory function has clearly confirmed the anecdotal reports of this phenomenon. subjective hyperosmia is associated with primarily negative odor-related experiences. Hyperosmia is increased olfactory acuity \textbackslash n Question: What is hyperosmia & \multirow{8}{*}{Hyperosmia is increased olfactory acuity.} \\
\midrule
\multirow{3}{*}{RE} & \multirow{3}{*}{\shortstack{Drug-Drug\\Interaction}} & Context: Antacids may interfere with the absorption of LEVSIN. Drug\_1: Antacids Drug\_2: LEVSIN & \multirow{3}{*}{\shortstack{true}} \\
\midrule
\multirow{3}{*}{\shortstack{Sentiment\\Analysis}} & \multirow{3}{*}{\shortstack{Medical\\Drugs}} & Why don't more folk opt for Cladribine? \textbackslash n Drug: cladribine \textbackslash n Option1: Neutral Option2: Positive Option3: Negative & \multirow{3}{*}{\shortstack{Positive}} \\
\midrule
\multirow{7}{*}{\shortstack{Systematic\\Review}} & \multirow{7}{*}{\shortstack{Acromegaly}} & No greater incidence or worsening of cardiac valve regurgitation with somatostatin analog treatment of acromegaly CONTEXT: Excess GH and IGF-I in acromegaly are associated with reduced life expectancy due to cardiovascular complications. Option\_1: Include, Option\_2: Exclude. & \multirow{7}{*}{Include} \\
\midrule
\multirow{3}{*}{\shortstack{Document\\Classification}} & \multirow{3}{*}{\shortstack{Hallmarks\\of Cancer\\(HoC)}} & Studies of cell-cycle progression showed that the anti-proliferative effect of Fan was associated with an increase in the G1/S phase of PC3 cells. & \multirow{3}{*}{\shortstack{Evading growth suppressors, Sustaining \\ proliferative signaling}} \\
\midrule
\multirow{8}{*}{\shortstack{Risk\\Factor\\Identification}} & \multirow{8}{*}{\shortstack{n2c2 - Risk \\ Factors Heart \\ Disease 2014 \\ (yesno)}} & Context: Record date: 2157-08-27 History of Present Illness ID:Admitted from cardiac cath lab. HPI:Mr. Doty is a 80 y.o. male with h/o HTN, DM, PVD, elevated cholesterol who presents with 6 month h/o chest and upper extremity discomfort on exertion along with SOB. He has limited his activities to prevent symptoms. ... \textbackslash n Risk Factor: Diabetes & \multirow{8}{*}{Yes} \\
\bottomrule
\end{tabular}
}
\caption{Examples of one instruction tasks converted into text-to-text format for each category}
\label{tab:examples}
\end{table*}

\section{Systematic Review Datasets}
\label{app:sr_datasets}

This section describes the brief data creation process for Systematic Reviews (SRs) that are used in this study. The relentless growth in clinical research and published articles have created a need for automation to expedite the process of SRs and to enable Living Systematic Reviews (LSRs). A crucial step in both SRs and LSRs is the title and abstract-based screening of the articles. A new dataset was developed from six SRs in the clinical domain by Mayo clinic physicians. In this study, we used data from the following five SRs that were conducted using the traditional (manual) process and published in relevant venues: (1) Hormone Replacement Therapy (HRT), (2) Cooking, (3) Accelerometer, (4) Acromegaly, and (5) COVID. The initial bibliographic search was designed and conducted by an experienced librarian with guidance from the principal investigators for the respective studies. The search was conducted in different bibliographic databases like PubMed, PubMed Central (PMC), Embase, EBM Reviews, and Ovid MEDLINE(R). Each article in the bibliographic search results was categorized by two physicians with domain expertise as ``Include'' or ``Exclude'', by reading the title and abstract of the article. When there was a disagreement between two annotators, a positive class (i.e., ``Include'') was preferred.

\section{Few-Shot Learning results}
\label{app:fewshot_results}

This section presents the results of few-shot learning for all instruction tasks in Table \ref{tab:fewshot-results}.

\begin{table*}[t]
\centering
\small
\resizebox{0.95\linewidth}{!}{

\begin{tabular}{@{}llcc|cc|cc|cc@{}}
\toprule
\multirow{2}{*}{Category} & \multirow{2}{*}{Task} &
\multicolumn{2}{c}{32} & 
\multicolumn{2}{c}{100} &  \multicolumn{2}{c}{$1k$} & \multicolumn{2}{c}{$4k$} \\
\cmidrule(lr){3-4}\cmidrule(lr){5-6}\cmidrule(lr){7-8}\cmidrule{9-10}
~& ~& S & I-BB & S & I-BB & S & I-BB & S & I-BB \\ 
 \toprule
 \multirow{16}{*}{NER}
 & AnatEM & 12.74 & 60.73 & 20.68 & 79.34 & 87.81 & 86.76 & 84.88 & 83.44 \\
 & BC2GM & 16.92& 65.65 & 21.31 & 70.39 & 82.92 & 77.19 & 77.66 & 74.11 \\
 & BC4CHEMD & 10.55 & 71.05 & 14.93 & 73.85 & 86.53 & 83.75 & 88.85 & 86.19 \\
 & BC5CDR & 11.75 & 60.37 & 12.58 & 67.51 & 69.62 & 73.66 & 74.83 & 74.34 \\
 & BioNLP11EPI & 31.14 & 78.64 & 42.31 & 81.51 & 85.71 & 85.57 & 84.64 & 86.68 \\
 & BioNLP11ID & 11.00 & 62.38 & 10.06 & 68.92 & 71.41 & 71.62 & 71.08 & 71.96 \\
 & BioNLP13CG & 12.39 & 49.15 & 12.53 & 52.68 & 55.23 & 63.15 & 64.19 & 67.23 \\
 & BioNLP13GE & 26.10 & 78.80 & 25.00 & 81.82 & 84.77 & 84.29 & 83.74 & 85.58 \\
 & BioNLP13PC & 12.40 & 69.29 & 12.59 & 71.89 & 68.11 & 68.49 & 70.42 & 71.97 \\
 & BioNLP09 & 32.51 & 78.17 & 30.51 & 82.71 & 87.48 & 86.39 & 85.16 & 86.33 \\
 & CRAFT & 8.07 & 37.35 & 8.60 & 40.38 & 49.67 & 51.56 & 63.72 & 63.35 \\
 & Ex-PTM & 16.06 & 74.32 & 47.93 & 76.15 & 82.92 & 84.11 & 82.32 & 83.81 \\
 & JNLPBA & 20.15 & 57.61 & 19.77 & 59.54 & 64.46 & 63.63 & 71.65 & 70.45 \\
 & NCBI & 38.69 & 68.82 & 30.46 & 79.35 & 93.02 & 90.36 & 89.51 & 86.46 \\
 & linnaeus & 28.75 & 58.69 & 36.94 & 67.29 & 93.81 & 92.50 & 94.43 & 70.57 \\
 & ---------------------------- & -------- & -------- & -------- & -------- & -------- & -------- & -------- & -------- \\
 & Average & 19.28 & \textbf{64.74} & 23.08 & \textbf{70.22} & 77.56 & 77.54 & \textbf{79.14} & 77.50\\
 \midrule
 \multirow{1}{*}{DI}
 & DI 2006 & 12.67 & \textbf{50.19} & 13.30 & \textbf{49.54} & 13.54 & \textbf{55.28} & 12.60 & \textbf{50.10} \\
 \midrule
 \multirow{1}{*}{POS}
 & Genia & \textbf{51.48} & 13.41 & \textbf{48.26} & 30.65 & \textbf{66.27} & 61.93 & \textbf{71.45} & 70.57 \\ 
 \midrule
 \multirow{2}{*}{QA}
 & BioASQ8b (factoid) & 36.63 & 35.99 & 41.89 & 40.77 & 51.96 & 49.84 & 52.95 & 51.72  \\
 & BioASQ8b (list) & 14.99 & 20.91 & 19.66 & 29.38 & 40.14 & 29.59 & 38.96 & 34.68 \\
 & BioASQ8b (yesno) & 43.48 & 61.11 & 39.13 & 57.94 & 66.96 & 60.32 & 56.52 & 52.17 \\
 & PubMedQA & 17.32 & 19.28 & 25.16 & 23.26 & 27.68 & 25.86 & 27.12 & 24.96 \\
 & ---------------------------- & -------- & -------- & -------- & -------- & -------- & -------- & -------- & -------- \\
 & Average & 28.11 & \textbf{34.32} & 31.46 & \textbf{37.84} & \textbf{46.68} & 41.40 & \textbf{43.89} & 40.88 \\ 
 \midrule
 \multirow{2}{*}{RE}
 & ChemProt & 61.64 & 72.02 & 66.07 & 64.91 & 66.01 & 55.22 & 76.86 & 77.38 \\
 & DDI & 85.53 & 77.37 & 85.53 & 81.37 & 46.99 & 55.41 & 87.39 & 73.04 \\
 & ---------------------------- & -------- & -------- & -------- & -------- & -------- & -------- & -------- & -------- \\
 & Average & 73.59 & \textbf{74.70} & \textbf{75.80} & 73.14 & \textbf{56.50} &	55.31 & \textbf{82.12} & 75.21 \\ 
 \midrule
 \multirow{1}{*}{SA}
 & Medical Drugs & 33.29 & \textbf{63.48} & 24.51 & \textbf{63.66} & \textbf{43.41} & 31.58 & 37.31 & \textbf{49.50} \\ 
 \midrule
 \multirow{6}{*}{SR}
 & Accelerometer & 76.76 & 77.78 & 75.35 & 68.06 & 83.80 & 73.61 & 72.54 & 70.83 \\
 & Acromegaly & 80.21 & 80.71 & 81.25 & 75.63 & 76.56 & 79.19 & 76.04 & 77.66 \\
 & COVID & 87.85 & 88.36 & 87.85 & 84.85 & 61.93 & 86.96 & 73.93 & 78.12 \\
 & Cooking & 88.29 & 87.08 & 87.80 & 87.56  & 81.95 & 87.08 & 80.98 & 82.78 \\
 & HRT & 85.86 & 86.02 & 85.61 & 75.12 & 89.08 & 81.99 & 83.87 & 80.81 \\
 & ---------------------------- & -------- & -------- & -------- & -------- & -------- & -------- & -------- & -------- \\
 & Average & 83.79 & \textbf{83.99} & \textbf{83.57} & 78.24 & 78.66 & \textbf{81.77} & 77.47 & \textbf{78.04} \\ 
 \midrule
 \multirow{1}{*}{DC}
 & HoC & 17.06 & \textbf{19.87} & 17.98 & \textbf{27.13} & 46.94 & \textbf{52.36} & \textbf{88.53} & 81.51 \\ 
 \midrule
 \multirow{3}{*}{RFI}
 & RFHD 2014 (yesno) & 57.21 & 51.78 & 57.21 & 51.50 & 43.02 & 66.35 & 43.86 & 66.46 \\
 & RFHD 2014 (time-riskfactor) & 54.51 & 64.22 & 52.75 & 63.37 & 66.18 & 59.60 & 66.18 & 62.70 \\
 & ---------------------------- & -------- & -------- & -------- & -------- & -------- & -------- & -------- & -------- \\
 & Average & 55.86 & \textbf{58.00} & 54.98 & \textbf{57.43} & 54.60 & \textbf{62.98} & 54.93 & \textbf{64.58} \\ 
 \midrule
 \multirow{1}{*}{Average}
 & - & 37.31 & \textbf{60.64} & 39.24 & \textbf{63.38} & 66.75 & \textbf{67.98} & 69.81 & \textbf{70.23} \\ 
 \bottomrule
\end{tabular}
}
\caption{Comparison of few-shot learning results in terms of Rouge-L between single-task models and In-BoXBART for 32/100/1000 training samples per instruction tasks. All results are presented in \%. S: Single-task model, I-BB: In-BoxBART, RFHD: Risk Factor for Heart Disease.}
\label{tab:fewshot-results}
\end{table*}

\section{State-of-the-art results}
\label{app:sota_results}

In Table \ref{tab:sota-results}, we present State-Of-The-Art (SOTA) results for 21 tasks. To compare the SOTA results with the proposed model, we calculate the corresponding metric used in particular research from our model predictions. For each task, we gather the best performance, and specifically, they are BioASQ-8b ~\citep{nentidis2020overview}, Chemprot ~\citep{peng2019transfer}, DDI ~\citep{peng2019transfer}. In Chemprot and DDI, we compare results with the base LMs instead of large for a fair comparison. SOTA results for all 15 NER datasets are obtained from \cite{banerjee2021biomedical}. Best performance for the HoC dataset is obtained from \cite{peng2019transfer}. Here, we have considered the result of the best system submitted to \cite{stubbs2015identifying} as SOTA result.

\begin{table*}[t]
\centering
\small
\resizebox{0.7\linewidth}{!}{

\begin{tabular}{@{}llcccc@{}}
\toprule
\multirow{2}{*}{Category} & \multirow{2}{*}{Task} & \multirow{2}{*}{Metric} & \multirow{2}{*}{SOTA} &  \multicolumn{2}{c}{Multi-Task} \\
\cmidrule(lr){5-6}
~& ~& ~ & ~ & V-BB & I-BB \\ 
 \toprule
 \multirow{16}{*}{NER}
 & AnatEM & F & 91.61 & 33.50 & 84.61 \\
 & BC2GM & F & 83.47 & 50.86 & 75.03 \\
 & BC4CHEMD & F & 92.39 & 71.44 & 86.97 \\
 & BC5CDR & F &  90.50  & 70.11 & 75.24 \\
 & BioNLP11EPI & F & 88.66 & 52.85 & 88.04 \\
 & BioNLP11ID & F & 87.36 & 60.15 & 73.39 \\
 & BioNLP13CG & F & 90.16 & 53.88 & 65.09 \\
 & BioNLP13GE & F & 85.81 & 51.78 & 87.39 \\
 & BioNLP13PC & F & 91.65 & 51.61 & 67.77 \\
 & BioNLP09 & F & 91.94 & 54.31 & 88.48 \\
 & CRAFT & F & 90.12 & 52.31 & 64.03 \\
 & Ex-PTM & F & 87.08 & 52.07 & 84.49 \\
 & JNLPBA & F & 79.19 & 68.60 & 70.26 \\
 & NCBI & F & 89.82 & 75.55 & 86.91 \\
 & linnaeus & F & 95.68 & 44.59 & 93.77 \\
 \midrule
 \multirow{2}{*}{QA}
 & BioASQ8 (list)& F & 52.99 & 17.74 & 35.59 \\
 & BioASQ8 (yesno)& F & 89.95 & 62.61 & 68.25 \\
 \midrule
 \multirow{2}{*}{RE}
 & Chemprot & F & 74.40 & 52.17 & 63.22 \\
 & DDI & F & 79.40 & 82.97 & 89.35 \\
 \midrule
 \multirow{1}{*}{DC}
 & HoC & F & 85.30 & 49.51 & 82.53 \\
 \midrule
 \multirow{1}{*}{RFI}
 & RFHD 2014 (time-riskfactor) & F & 92.76 & 0.97 & 85.28 \\
 \midrule
 \multirow{1}{*}{Average}
 & - & - & 85.55 & 50.36 & 72.24 \\
 \bottomrule
\end{tabular}
}
\caption{The state-of-the-art (SOTA) results for each task compared with Vanilla-BoXBART and In-BoXBART. All results are in \%. F: $F_1$-score, V-BB: Vanilla-BoXBART, I-BB: In-BoXBART, RFHD: Risk Factor for Heart Disease.}
\label{tab:sota-results}
\end{table*}

\end{document}